# MULTI-PARAGRAPH SEGMENTATION OF EXPOSITORY TEXT

Marti A. Hearst
Computer Science Division, 571 Evans Hall
University of California, Berkeley
Berkeley, CA 94720
and
Xerox Palo Alto Research Center
*marti@cs.berkeley.edu*

## Abstract

This paper describes TextTiling, an algorithm for partitioning expository texts into coherent multi-paragraph discourse units which reflect the subtopic structure of the texts. The algorithm uses domain-independent lexical frequency and distribution information to recognize the interactions of multiple simultaneous themes. Two fully-implemented versions of the algorithm are described and shown to produce segmentation that corresponds well to human judgments of the major subtopic boundaries of thirteen lengthy texts.

## INTRODUCTION

The structure of expository texts can be characterized as a sequence of subtopical discussions that occur in the context of a few main topic discussions. For example, a popular science text called *Stargazers*, whose main topic is the existence of life on earth and other planets, can be described as consisting of the following subdiscussions (numbers indicate paragraph numbers):

|       |                                                      |
|-------|------------------------------------------------------|
| 1-3   | Intro – the search for life in space                 |
| 4-5   | The moon's chemical composition                      |
| 6-8   | How early proximity of the moon shaped it            |
| 9-12  | How the moon helped life evolve on earth             |
| 13    | Improbability of the earth-moon system               |
| 14-16 | Binary/trinary star systems make life unlikely       |
| 17-18 | The low probability of non-binary/trinary systems    |
| 19-20 | Properties of our sun that facilitate life           |
| 21    | Summary                                              |

Subtopic structure is sometimes marked in technical texts by headings and subheadings which divide the text into coherent segments; Brown & Yule (1983:140) state that this kind of division is one of the most basic in discourse. However, many expository texts consist of long sequences of paragraphs with very little structural demarcation. This paper presents fully-implemented algorithms that use lexical cohesion relations to partition expository texts into multi-paragraph segments that reflect their subtopic structure. Because the model of discourse structure is one in which text is partitioned into contiguous, nonoverlapping blocks, I call the general approach TextTiling. The ultimate goal is to not only identify the extents of the subtopical units, but to label their contents as well. This paper focusses only on the discovery of subtopic structure, leaving determination of subtopic content to future work.

Most discourse segmentation work is done at a finer granularity than that suggested here. However, for lengthy written expository texts, multi-paragraph segmentation has many potential uses, including the improvement of computational tasks that make use of distributional information. For example, disambiguation algorithms that train on arbitrary-size text windows, e.g., Yarowsky (1992) and Gale *et al.* (1992b), and algorithms that use lexical co-occurrence to determine semantic relatedness, e.g., Schütze (1993), might benefit from using windows with motivated boundaries instead.

Information retrieval algorithms can use subtopic structuring to return meaningful portions of a text if paragraphs are too short and sections are too long (or are not present). Motivated segments can also be used as a more meaningful unit for indexing long texts. Salton *et al.* (1993), working with encyclopedia text, find that comparing a query against sections and then paragraphs is more successful than comparing against full documents alone. I have used the results of Text-Tiling in a new paradigm for information access on full-text documents (Hearst 1994).

The next section describes the discourse model that motivates the approach. This is followed by a description of two algorithms for subtopic structuring that make use only of lexical cohesion relations, the evaluation of these algorithms, and a summary and discussion of future work.

## THE DISCOURSE MODEL

Many discourse models assume a hierarchical segmentation model, e.g., attentional/intentional structure (Grosz & Sidner 1986) and Rhetorical Structure Theory (Mann & Thompson 1987). Although many aspects of discourse analysis require such a model, I choose to cast expository text into a linear sequence of segments, both for computational simplicity and because such a structure is sufficient for the coarse-grained tasks of interest here.[1]

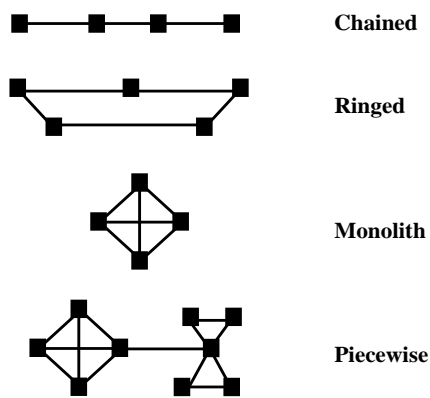

Figure 1: Skorochod'ko's text structure types. Nodes correspond to sentences and edges between nodes indicate strong term overlap between the sentences.

Skorochod'ko (1972) suggests discovering a text's structure by dividing it up into sentences and seeing how much word overlap appears among the sentences. The overlap forms a kind of intra-structure; fully connected graphs might indicate dense discussions of a topic, while long spindly chains of connectivity might indicate a sequential account (see Figure 1). The central idea is that of defining the structure of a text as a function of the connectivity patterns of the terms that comprise it. This is in contrast with segmenting guided primarily by fine-grained discourse cues such as register change, focus shift, and cue words. From a computational viewpoint, deducing textual topic structure from lexical connectivity alone is appealing, both because it is easy to compute, and also because discourse cues are sometimes misleading with respect to the topic structure (Brown & Yule 1983)(§3).

The topology most of interest to this work is the final one in the diagram, the Piecewise Monolithic Structure, since it represents sequences of densely interrelated discussions linked together, one after another. This topology maps nicely onto that of viewing documents as a sequence of densely interrelated subtopical discussions, one following another. This assumption, as will be seen, is not always valid, but is nevertheless quite useful.

This theoretical stance bears a close resemblance to Chafe's notion of The Flow Model of discourse (Chafe 1979), in description of which he writes (pp 179-180):

> Our data ... suggest that as a speaker moves from focus to focus (or from thought to thought) there are certain points at which there may be a more or less radical change in space, time, character configuration, event structure, or, even, world. ... At points where all of these change in a maximal way, an episode boundary is strongly present. But often one or another will change considerably while others will change less radically, and all kinds of varied interactions between these several factors are possible.[2]

Although Chafe's work concerns narrative text, the same kind of observation applies to expository text. The TextTiling algorithms are designed to recognize episode boundaries by determining where thematic components like those listed by Chafe change in a maximal way.

Many researchers have studied the patterns of occurrence of characters, setting, time, and the other thematic factors that Chafe mentions, usually in the context of narrative. In contrast, I attempt to determine where a relatively large set of active themes changes simultaneously, regardless of the *type* of thematic factor. This is especially important in expository text in which the subject matter tends to structure the discourse more so than characters, setting, etc. For example, in the *Stargazers* text, a discussion of continental movement, shoreline acreage, and habitability gives way to a discussion of binary and unary star systems. This is not so much a change in setting or character as a change in subject matter. Therefore, to recognize where the subtopic changes occur, I make use of lexical cohesion relations (Halliday & Hasan 1976) in a manner similar to that suggested by Skorochod'ko.

Morris and Hirst's pioneering work on computing discourse structure from lexical relations (Morris & Hirst 1991), (Morris 1988) is a precursor to the work reported on here. Influenced by Halliday & Hasan's (1976) theory of lexical coherence, Morris developed an algorithm that finds chains of related terms via a comprehensive thesaurus (Roget's Fourth Edition).[3] For example, the

---

[1] Additionally, (Passonneau & Litman 1993) concede the difficulty of eliciting hierarchical intentional structure with any degree of consistency from their human judges.

[2] Interestingly, Chafe arrived at the Flow Model after working extensively with, and then becoming dissatisfied with, a hierarchical model of paragraph structure like that of Longacre (1979).

[3] The algorithm is executed by hand since the thesaurus

```
     Sentence:     05    10    15    20    25    30    35    40    45    50    55    60    65    70    75    80    85    90    95
    14      form   1      111 1     1                                    1 1    1     1         1     1       1     1
     8 scientist                    11              1     1                     1           1         1 1  1
     5     space  11   1        1                                                                 1
    25      star   1             1                                            11 22   111112  1 1   1    11 1111     1
     5    binary                                                              11  1                 1                1
     4   trinary                                                               1   1                1                1
     8 astronomer  1             1                                             1 1              1   1    1 1
     7     orbit   1                      1                                        12       1 1
     6      pull                             2        1 1                                   1 1
    16    planet   1    1        11              1            1                    21  11111                    1    1
     7    galaxy   1                                                       1              1 11       1                1
     4     lunar              1 1      1          1
    19      life   1 1 1                                 1     11 1   11 1    1                     1 1        1 111 1 1
    27      moon       13    1111    1 1 22 21    21       21               11 1
     3      move                                              1  1  1
     7 continent                                            2 1 1 2 1
     3 shoreline                                                12
     6      time                           1                 1 1 1     1                                                   1
     3     water                                   11                  1
     6       say                                   1 1         1          11                   1
     3   species                                            1 1 1
     Sentence:     05    10    15    20    25    30    35    40    45    50    55    60    65    70    75    80    85    90    95
```

Figure 2: Distribution of selected terms from the *Stargazer* text, with a single digit frequency per sentence number (blanks indicate a frequency of zero).

words *residential* and *apartment* both index the same thesaural category and can thus be considered to be in a coherence relation with one another. The chains are used to structure texts according to the attentional/intentional theory of discourse structure (Grosz & Sidner 1986), and the extent of the chains correspond to the extent of a segment. The algorithm also incorporates the notion of "chain returns" – repetition of terms after a long hiatus – to close off an intention that spans over a digression.

Since the Morris & Hirst (1991) algorithm attempts to discover attentional/intentional structure, their goals are different than those of TextTiling. Specifically, the discourse structure they attempt to discover is hierarchical and more fine-grained than that discussed here. Thus their model is not set up to take advantage of the fact that multiple simultaneous chains might occur over the same intention. Furthermore, chains tend to overlap one another extensively in long texts. Figure 2 shows the distribution, by sentence number, of selected terms from the *Stargazers* text. The first two terms have fairly uniform distribution and so should not be expected to provide much information about the divisions of the discussion. The next two terms occur mainly at the beginning and the end of the text, while terms *binary* through *planet* have considerable overlap from sentences 58 to 78. There is a somewhat well-demarked cluster of terms between sentences 35 and 50, corresponding to the grouping together of paragraphs 10, 11, and 12 by human judges who have read the text.

From the diagram it is evident that simply looking for chains of repeated terms is not sufficient for determining subtopic breaks. Even combining terms that are closely related semantically into single chains is insufficient, since often several different themes are active in the same segment. For example, sentences 37 - 51 contain dense interaction among the terms *move*, *continent*, *shoreline*, *time*, *species*, and *life*, and all but the latter occur only in this region. However, it is the case that the interlinked terms of sentences 57 - 71 (*space*, *star*, *binary*, *trinary*, *astronomer*, *orbit*) are closely related semantically, assuming the appropriate senses of the terms have been determined.

## ALGORITHMS FOR DISCOVERING SUBTOPIC STRUCTURE

Many researchers (e.g., Halliday & Hasan (1976), Tannen (1989), Walker (1991)) have noted that term repetition is a strong cohesion indicator. I have found in this work that term repetition alone is a very useful indicator of subtopic structure, when analyzed in terms of multiple simultaneous information threads. This section describes two algorithms for discovering subtopic is not generally available online.

structure using term repetition as a lexical cohesion indicator.

The first method compares, for a given window size, each pair of adjacent blocks of text according to how similar they are lexically. This method assumes that the more similar two blocks of text are, the more likely it is that the current subtopic continues, and, conversely, if two adjacent blocks of text are dissimilar, this implies a change in subtopic flow. The second method, an extension of Morris & Hirst's (1991) approach, keeps track of active chains of repeated terms, where membership in a chain is determined by location in the text. The method determines subtopic flow by recording where in the discourse the bulk of one set of chains ends and a new set of chains begins.

The core algorithm has three main parts:

1. Tokenization
2. Similarity Determination
3. Boundary Identification

Tokenization refers to the division of the input text into individual lexical units. For both versions of the algorithm, the text is subdivided into psuedosentences of a pre-defined size $w$ (a parameter of the algorithm) rather than actual syntactically-determined sentences, thus circumventing normalization problems. For the purposes of the rest of the discussion these groupings of tokens will be referred to as *token-sequences*. In practice, setting $w$ to 20 tokens per token-sequence works best for many texts. The morphologically-analyzed token is stored in a table along with a record of the token-sequence number it occurred in, and how frequently it appeared in the token-sequence. A record is also kept of the locations of the paragraph breaks within the text. Closed-class and other very frequent words are eliminated from the analysis.

After tokenization, the next step is the comparison of adjacent pairs of blocks of token-sequences for overall lexical similarity. Another important parameter for the algorithm is the *blocksize*: the number of token-sequences that are grouped together into a block to be compared against an adjacent group of token-sequences. This value, labeled $k$, varies slightly from text to text; as a heuristic it is the average paragraph length (in token-sequences). In practice, a value of $k = 6$ works well for many texts. Actual paragraphs are not used because their lengths can be highly irregular, leading to unbalanced comparisons.

Similarity values are computed for every token-sequence gap number; that is, a score is assigned to token-sequence gap $i$ corresponding to how similar the token-sequences from token-sequence $i-k$ through $i$ are to the token-sequences from $i+1$ to $i+k+1$. Note that this moving window approach means that each token-sequence appears in $k*2$ similarity computations.

Similarity between blocks is calculated by a cosine measure: given two text blocks $b_1$ and $b_2$, each with $k$ token-sequences,

$$sim(b_1, b_2) = \frac{\sum_t w_{t,b_1} w_{t,b_2}}{\sqrt{\sum_t w_{t,b_1}^2 \sum_{t=1}^n w_{t,b_2}^2}}$$

where $t$ ranges over all the terms that have been registered during the tokenization step, and $w_{t,b_1}$ is the weight assigned to term $t$ in block $b_1$. In this version of the algorithm, the weights on the terms are simply their frequency within the block.[4] Thus if the similarity score between two blocks is high, then the blocks have many terms in common. This formula yields a score between 0 and 1, inclusive.

These scores can be plotted, token-sequence number against similarity score. However, since similarity is measured between blocks $b_1$ and $b_2$, where $b_1$ spans token-sequences $i - k$ through $i$ and $b_2$ spans $i + 1$ to $i + k + 1$, the measurement's $x$-axis coordinate falls between token-sequences $i$ and $i + 1$. Rather than plotting a token-sequence number on the $x$-axis, we plot token-sequence *gap* number $i$. The plot is smoothed with average smoothing; in practice one round of average smoothing with a window size of three works best for most texts.

Boundaries are determined by changes in the sequence of similarity scores. The token-sequence gap numbers are ordered according to how steeply the slopes of the plot are to either side of the token-sequence gap, rather than by their absolute similarity score. For a given token-sequence gap $i$, the algorithm looks at the scores of the token-sequence gaps to the left of $i$ as long are their values are increasing. When the values to the left peak out, the difference between the score at the peak and the score at $i$ is recorded. The same procedure takes place with the token-sequence gaps to the right of $i$; their scores are examined as long as they continue to rise. The relative height of the peak to the right of $i$ is added to the relative height of the peak to the left. (A gap occurring at a peak will have a score of zero since neither of its neighbors is higher than it.) These new scores, called depth scores, corresponding to how sharp a change occurs on both sides of the token-sequence gap, are then sorted. Segment boundaries are assigned to the token-sequence gaps with the largest corresponding scores, adjusted as necessary to correspond to true paragraph breaks. A proviso check is done that prevents assignment of very close adjacent segment boundaries. Currently there must be at least three intervening token-sequences between boundaries. This helps control for the fact that many texts have spurious header information and single-sentence paragraphs.

The algorithm must determine how many segments to assigned to a document, since every paragraph is a potential segment boundary. Any attempt to make an absolute cutoff is problematic since there would need to be some correspondence to the document style and length. A cutoff based on a particular valley depth is similarly problematic.

---

[4]Earlier work weighted the terms according to their frequency times their inverse document frequency. In these more recent experiments, simple term frequencies seem to work better.

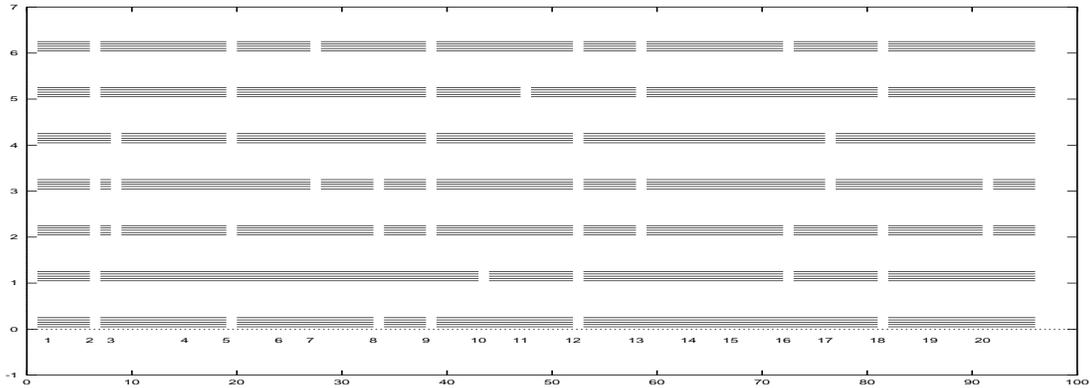

Figure 3: Judgments of seven readers on the *Stargazer* text. Internal numbers indicate location of gaps between paragraphs; x-axis indicates token-sequence gap number, y-axis indicates judge number, a break in a horizontal line indicates a judge-specified segment break.

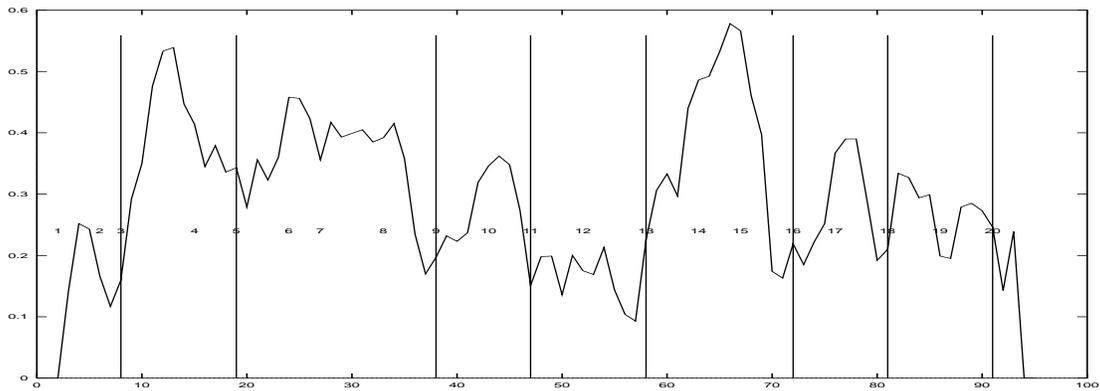

Figure 4: Results of the block similarity algorithm on the *Stargazer* text. Internal numbers indicate paragraph numbers, x-axis indicates token-sequence gap number, y-axis indicates similarity between blocks centered at the corresponding token-sequence gap. Vertical lines indicate boundaries chosen by the algorithm; for example, the leftmost vertical line represents a boundary after paragraph 3. Note how these align with the boundary gaps of Figure 3 above.

I have devised a method for determining the number of boundaries to assign that scales with the size of the document and is sensitive to the patterns of similarity scores that it produces: the cutoff is a function of the average and standard deviations of the depth scores for the text under analysis. Currently a boundary is drawn only if the depth score exceeds $\bar{s} - \sigma/2$.

## EVALUATION

One way to evaluate these segmentation algorithms is to compare against judgments made by human readers, another is to compare the algorithms against texts pre-marked by authors, and a third way is to see how well the results improve a computational task. This section compares the algorithm against reader judgments, since author markups are fallible and are usually applied to text types that this algorithm is not designed for, and Hearst (1994) shows how to use TextTiles in a task (although it does not show whether or not the results of the algorithms used here are better than some other algorithm with similar goals).

### Reader Judgments

Judgments were obtained from seven readers for each of thirteen magazine articles which satisfied the length criteria (between 1800 and 2500 words)[5] and which contained little structural demarkation. The judges were asked simply to mark the paragraph boundaries at which the topic changed; they were not given more explicit instructions about the granularity of the segmentation.

---

[5] One longer text of 2932 words was used since reader judgments had been obtained for it from an earlier experiment. Judges were technical researchers. Two texts had three or four short headers which were removed for consistency.

Figure 3 shows the boundaries marked by seven judges on the *Stargazers* text. This format helps illustrate the general trends made by the judges and also helps show where and how often they disagree. For instance, all but one judge marked a boundary between paragraphs 2 and 3. The dissenting judge did mark a boundary after 3, as did two of the concurring judges. The next three major boundaries occur after paragraphs 5, 9, 12, and 13. There is some contention in the later paragraphs; three readers marked both 16 and 18, two marked 18 alone, and two marked 17 alone. The outline in the Introduction gives an idea of what each segment is about.

Passonneau & Litman (1993) discuss at length considerations about evaluating segmentation algorithms according to reader judgment information. As Figure 3 shows, agreement among judges is imperfect, but trends can be discerned. In Passonneau & Litman's (1993) data, if 4 or more out of 7 judges mark a boundary, the segmentation is found to be significant using a variation of the Q-test (Cochran 1950). My data showed similar results. However, it isn't clear how useful this significance information is, since a simple majority does not provide overwhelming proof about the objective reality of the subtopic break. Since readers often disagree about where to draw a boundary marking for a topic shift, one can only use the general trends as a basis from which to compare different algorithms. Since the goals of TextTiling are better served by algorithms that produce more rather than fewer boundaries, I set the cutoff for "true" boundaries to three rather than four judges per paragraph.[6] The remaining gaps are considered nonboundaries.

### Results

Figure 4 shows a plot of the results of applying the block comparison algorithm to the *Stargazer* text. When the lowermost portion of a valley is not located at a paragraph gap, the judgment is moved to the nearest paragraph gap.[7] For the most part, the regions of strong similarity correspond to the regions of strong agreement among the readers. (The results for this text were fifth highest out of the 13 test texts.) Note however, that the similarity information around paragraph 12 is weak. This paragraph briefly summarizes the contents of the previous three paragraphs; much of the terminology that occurred in all of them reappears in this one location (in the spirit of a Grosz & Sidner (1986) "pop" operation). Thus it displays low similarity both to itself and to its neighbors. This is an example of a breakdown caused by the assumptions about the subtopic structure. It is possible that an additional pass through the text could be used to find structure of this kind.

The final paragraph is a summary of the entire text; the algorithm recognizes the change in terminology from the preceding paragraphs and marks a boundary; only two of the readers chose to differentiate the summary; for this reason the algorithm is judged to have made an error even though this sectioning decision is reasonable. This illustrates the inherent fallibility of testing against reader judgments, although in part this is because the judges were given loose constraints.

Following the advice of Gale *et al.* (1992a), I compare the algorithm against both upper and lower bounds. The upper bound in this case is the reader judgment data. The lower bound is a baseline algorithm that is a simple, reasonable approach to the problem that can be automated. A simple way to segment the texts is to place boundaries randomly in the document, constraining the number of boundaries to equal that of the average number of paragraph gaps assigned by judges. In the test data, boundaries are placed in about 41% of the paragraph gaps. A program was written that places a boundary at each potential gap 41% of the time (using a random number generator), and run 10,000 times for each text, and the average of the scores of these runs was found. These scores appear in Table 1 (results at 33% are also shown for comparison purposes).

The algorithms are evaluated according to how many true boundaries they select out of the total selected (precision) and how many true boundaries are found out of the total possible (recall) (Salton 1988). The recall measure implicitly signals the number of missed boundaries (false negatives, or deletion errors); the number of false positives, or insertion errors, is indicated explicitly.

In many cases the algorithms are almost correct but off by one paragraph, especially in the texts that the algorithm performs poorly on. When the block similarity algorithm is allowed to be off by one paragraph, there is dramatic improvement in the scores for the texts that lower part of Table 2, yielding an overall precision of 83% and recall of 78%. As in Figure 4, it is often the case that where the algorithm is incorrect, e.g., paragraph gap 11, the overall blocking is very close to what the judges intended.

Table 1 shows that both the blocking algorithm and the chaining algorithm are sandwiched between the upper and lower bounds. Table 2 shows some of these results in more detail. The block similarity algorithm seems to work slightly better than the chaining algorithm, although the difference may not prove significant over the long run. Furthermore, in both versions of the algorithm, changes to the parameters of the algorithm perturbs the resulting boundary markings. This is an undesirable property and perhaps could be remedied with some kind of information-theoretic formulation of the problem.

---

[6]Paragraphs of three or fewer sentences were combined with their neighbor if that neighbor was deemed to follow at "true" boundary, as in paragraphs 2 and 3 of the *Stargazers* text.

[7]This might be explained in part by (Stark 1988) who shows that readers disagree measurably about where to place paragraph boundaries when presented with texts with those boundaries removed.

|  | Precision | | Recall | |
| --- | --- | --- | --- | --- |
|  | $\bar{s}$ | $\sigma$ | $\bar{s}$ | $\sigma$ |
| Baseline 33% | .44 | .08 | .37 | .04 |
| Baseline 41% | .43 | .08 | .42 | .03 |
| Chains | .64 | .17 | .58 | .17 |
| Blocks | .66 | .18 | .61 | .13 |
| Judges | .81 | .06 | .71 | .06 |

Table 1: Precision and Recall values for 13 test texts.

## SUMMARY AND FUTURE WORK

This paper has described algorithms for the segmentation of expository texts into discourse units that reflect the subtopic structure of expository text. I have introduced the notion of the recognition of multiple simultaneous themes, which bears some resemblance to Chafe's Flow Model of discourse and Skorochod'ko's text structure types. The algorithms are fully implemented: term repetition alone, without use of thesaural relations, knowledge bases, or inference mechanisms, works well for many of the experimental texts. The structure it obtains is coarse-grained but generally reflects human judgment data.

Earlier work (Hearst 1993) incorporated thesaural information into the algorithms; surprisingly the latest experiments find that this information degrades the performance. This could very well be due to problems with the algorithm used. A simple algorithm that just posits relations among terms that are a small distance apart according to WordNet (Miller *et al.* 1990) or Roget's 1911 thesaurus (from Project Gutenberg), modeled after Morris and Hirst's heuristics, might work better. Therefore I do not feel the issue is closed, and instead consider successful grouping of related words as future work. As another possible alternative Kozima (1993) has suggested using a (computationally expensive) semantic similarity metric to find similarity among terms within a small window of text (5 to 7 words). This work does not incorporate the notion of multiple simultaneous themes but instead just tries to find breaks in semantic similarity among a small number of terms. A good strategy may be to substitute this kind of similarity information for term repetition in algorithms like those described here. Another possibility would be to use semantic similarity information as computed in Schütze (1993), Resnik (1993), or Dagan *et al.* (1993).

The use of discourse cues for detection of segment boundaries and other discourse purposes has been extensively researched, although predominantly on spoken text (see Hirschberg & Litman (1993) for a summary of six research groups' treatments of 64 cue words). It is possible that incorporation of such information may provide a relatively simple way improve the cases where the algorithm is off by one paragraph.

## Acknowledgments

This paper has benefited from the comments of Graeme Hirst, Jan Pedersen, Penni Sibun, and Jeff Siskind. I would like to thank Anne Fontaine for her interest and help in the early stages of this work, and Robert Wilensky for supporting this line of research. This work was sponsored in part by the Advanced Research Projects Agency under Grant No. MDA972-92-J-1029 with the Corporation for National Research Initiatives (CNRI), and by the Xerox Palo Alto Research Center.

| Text | Total Possible | Baseline 41% (avg) | | | | Blocks | | | | Chains | | | | Judges (avg) | | | |
|---|---|---|---|---|---|---|---|---|---|---|---|---|---|---|---|---|---|
| | | Prec | Rec | C | I | Prec | Rec | C | I | Prec | Rec | C | I | Prec | Rec | C | I |
| 1 | 9 | .44 | .44 | 4 | 5 | 1.0 | .78 | 7 | 0 | 1.0 | .78 | 7 | 0 | .78 | .78 | 7 | 2 |
| 2 | 9 | .50 | .44 | 4 | 4 | .88 | .78 | 7 | 1 | .75 | .33 | 3 | 1 | .88 | .78 | 7 | 1 |
| 3 | 9 | .40 | .44 | 4 | 6 | .78 | .78 | 7 | 2 | .56 | .56 | 5 | 4 | .75 | .67 | 6 | 2 |
| 4 | 12 | .63 | .42 | 5 | 3 | .86 | .50 | 6 | 1 | .56 | .42 | 5 | 4 | .91 | .83 | 10 | 1 |
| 5 | 8 | .43 | .38 | 3 | 4 | .70 | .75 | 6 | 2 | .86 | .75 | 6 | 1 | .86 | .75 | 6 | 1 |
| 6 | 8 | .40 | .38 | 3 | 9 | .60 | .75 | 6 | 3 | .42 | .63 | 5 | 8 | .75 | .75 | 6 | 2 |
| 7 | 9 | .36 | .44 | 4 | 7 | .60 | .56 | 5 | 3 | .40 | .44 | 4 | 6 | .75 | .67 | 6 | 2 |
| 8 | 8 | .43 | .38 | 3 | 4 | .50 | .63 | 5 | 4 | .67 | .75 | 6 | 3 | .86 | .75 | 6 | 1 |
| 9 | 9 | .36 | .44 | 4 | 7 | .50 | .44 | 4 | 3 | .60 | .33 | 3 | 2 | .75 | .67 | 6 | 2 |
| 10 | 8 | .50 | .38 | 3 | 3 | .50 | .50 | 4 | 3 | .63 | .63 | 5 | 3 | .86 | .75 | 6 | 1 |
| 11 | 9 | .36 | .44 | 4 | 7 | .50 | .44 | 4 | 4 | .71 | .56 | 5 | 2 | .75 | .67 | 6 | 2 |
| 12 | 9 | .44 | .44 | 4 | 5 | .50 | .56 | 5 | 5 | .54 | .78 | 7 | 6 | .86 | .67 | 6 | 1 |
| 13 | 10 | .36 | .40 | 4 | 7 | .30 | .50 | 5 | 9 | .60 | .60 | 6 | 4 | .78 | .70 | 7 | 2 |

Table 2: Scores by text, showing precision and recall. (C) indicates the number of correctly placed boundaries, (I) indicates the number of inserted boundaries. The number of deleted boundaries can be determined by subtracting (C) from Total Possible.